%% file: TECS.tex
\documentclass{article}

\usepackage{arxiv}

\usepackage{longtable}
\usepackage{xcolor, soul}
\usepackage{amsmath,amssymb,amsfonts}
\usepackage{algorithmic}
\usepackage{graphicx, subfigure}
\usepackage{mathtools}
\usepackage{textcomp}
\usepackage{xcolor}
\usepackage{booktabs,siunitx}
\usepackage{adjustbox}
\usepackage{multirow}
\usepackage{hyperref}
\sethlcolor{green}

\input{Tex/TitleAndAuthors}

\begin{document}
\maketitle


\input{Tex/Abstract}

\input{Tex/Intro}

\input{Tex/RelatedWork}

\input{Tex/Background}

\input{Tex/Architecture}

\input{Tex/ExpResults}

\input{Tex/Conclusion}


\bibliographystyle{IEEEtran}
\bibliography{Bilbo/Ref}

\clearpage
\input{Tex/Appendix}

\end{document}

%% file: Tex/TitleAndAuthors.tex
\title{ATCN: Resource-Efficient Processing of Time Series on Edge}

\author{
    Mohammadreza Baharani
    \thanks{Published in ACM Transactions on Embedded Computing Systems. Available at: \href{https://dl.acm.org/doi/10.1145/3524070}{https://dl.acm.org/doi/10.1145/3524070}}\\
    The UNC at Charlotte\\
    Charlotte, NC\\
    \texttt{mbaharan@uncc.edu}\\
    \And
    Hamed Tabkhi\\
    The UNC at Charlotte\\
    Charlotte, NC\\
    \texttt{htabkhiv@uncc.edu}\\
    }

%% file: Tex/Abstract.tex
\begin{abstract}
This paper presents a scalable deep learning model called Agile Temporal Convolutional Network (ATCN) for high-accurate fast classification and time series prediction in resource-constrained embedded systems. ATCN is a family of compact networks with formalized hyperparameters that enable application-specific adjustments to be made to the model architecture. It is primarily designed for embedded edge devices with very limited performance and memory, such as wearable biomedical devices and real-time reliability monitoring systems. ATCN makes fundamental improvements over the mainstream temporal convolutional neural networks, including residual connections to increase the network depth and accuracy, and the incorporation of separable depth-wise convolution to reduce the computational complexity of the model.  As part of the present work, two ATCN families, namely T0, and T1 are also presented and evaluated on different ranges of embedded processors - Cortex-M7 and Cortex-A57 processor. An evaluation of the ATCN models against the best-in-class InceptionTime and MiniRocket shows that ATCN almost maintains accuracy while improving the execution time on a broad range of embedded and cyber-physical applications with demand for real-time processing on the embedded edge. At the same time, in contrast to existing solutions, ATCN is the first time-series classifier based on deep learning that can be run bare-metal on embedded microcontrollers (Cortex-M7) with limited computational performance and memory capacity while delivering state-of-the-art accuracy.
\end{abstract}

\keywords{Temporal Convolutional Networks (TCN),  Recurrent Neural Networks (RNN), Real-time Edge Computing}

\maketitle

%% file: Tex/Intro.tex
\section{Introduction}
The rapid growth in deep learning algorithms has changed how embedded and cyber-physical systems (CPS) process the surrounding environment and has significantly improved the overall CPS performance on delivering their assigned tasks.  Time series analysis and forecasting are important applications that can significantly benefit from the deep learning paradigm on a broad range of smart IoT and embedded edge devices. Few pronounced examples are wearable health monitoring devices \cite{ecg_tcn, health_1, health_2}, on-board predictive maintenance \cite{Deep_race, biglar_dev_health, dev_lstm_3}, machine translation \cite{lstm_machine_trans1, gru_mt}, and smart transportation systems \cite{deo2018convolutional, xie2021congestion, mercat2020multi}. These applications all demand real-time accurate processing of time series on edge devices with limited computational resources.

For most deep learning practitioners, recurrent networks and especially two elaborated models, namely, LSTM \cite{lstm_main} and GRU \cite{chung2014empirical}, are synonymous with time series analysis due to its notable success in sequence modeling problems such as machine translation, language processing, and device health monitoring. These models interpolate the output based on the current and temporal information, which is learned and propagated across the hidden states sequentially. The propagation chain of hidden states causes two major issues \cite{DBLP:empirical}: 1) gradient instability such as vanishing/exploding gradients and 2) fewer levels of parallelization due to existing dependencies across the cells.

\begin{figure*}[t]
	\centering
	
	\subfigure[T0 and MiniRocket pairwise accuracy comparison]{%
		{\includegraphics[width=.45\linewidth, trim= 10 5 45 25,clip, keepaspectratio]{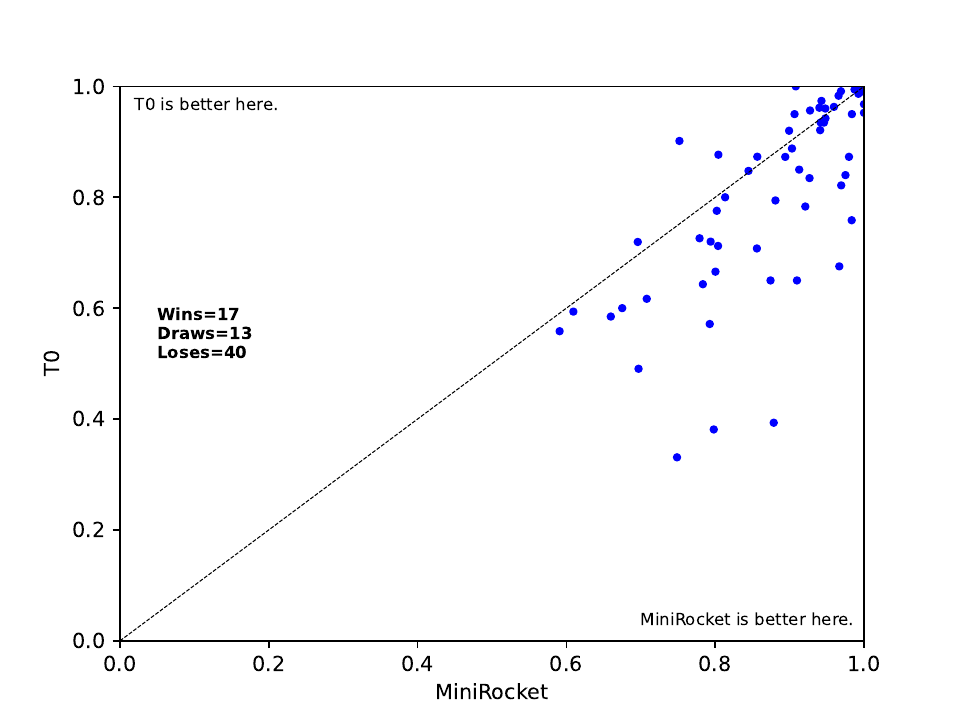}}%
		\label{fig:t0_mini_acc}%
	}\qquad
 	\subfigure[T0 and MiniRocket pairwise latency comparison. The unit of measurement is second.]{%
 		{\includegraphics[width=0.45\textwidth, trim= 10 5 45 25,clip, keepaspectratio]{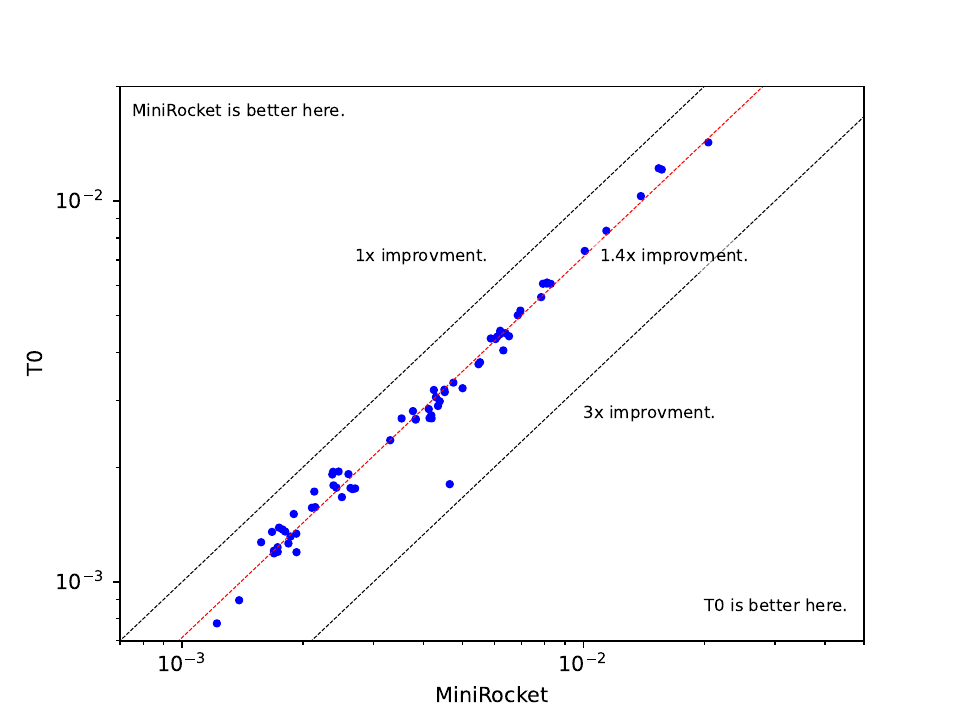}}%
 		\label{fig:t0_mini_latency}%
 	}\\
 	\subfigure[T1 and MiniRocket pairwise accuracy comparison.]{%
		{\includegraphics[width=.45\linewidth, trim= 10 5 45 25,clip, keepaspectratio]{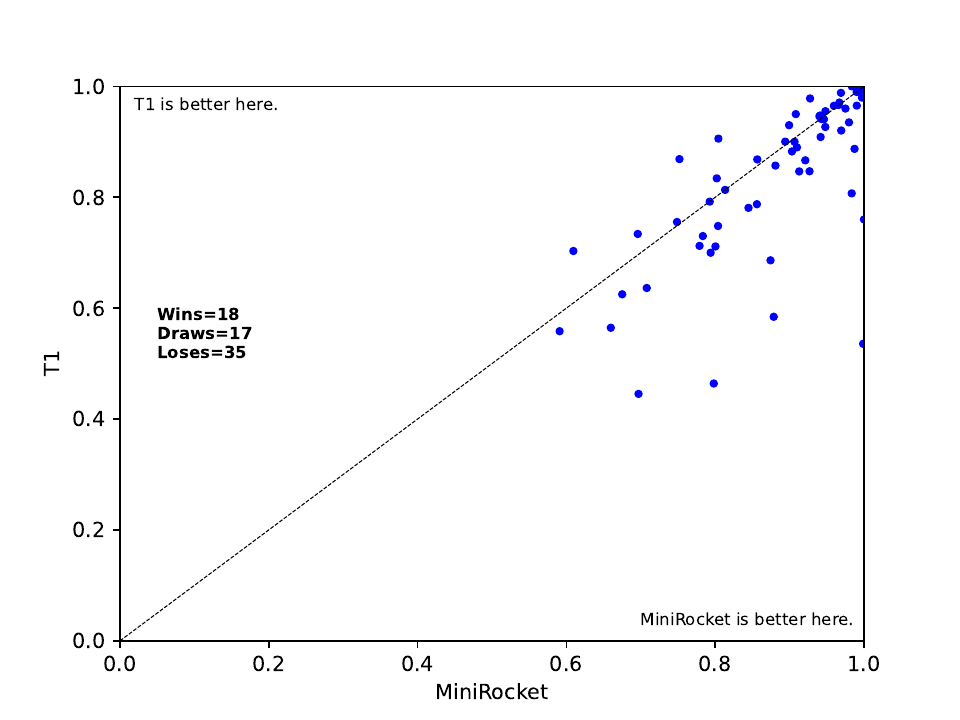}}%
		\label{fig:t1_mini_acc}%
	}\qquad
 	\subfigure[T1 and MiniRocket pairwise latency comparison. The unit of measurement is second.]{%
 		{\includegraphics[width=0.45\textwidth, trim= 10 5 45 25,clip, keepaspectratio]{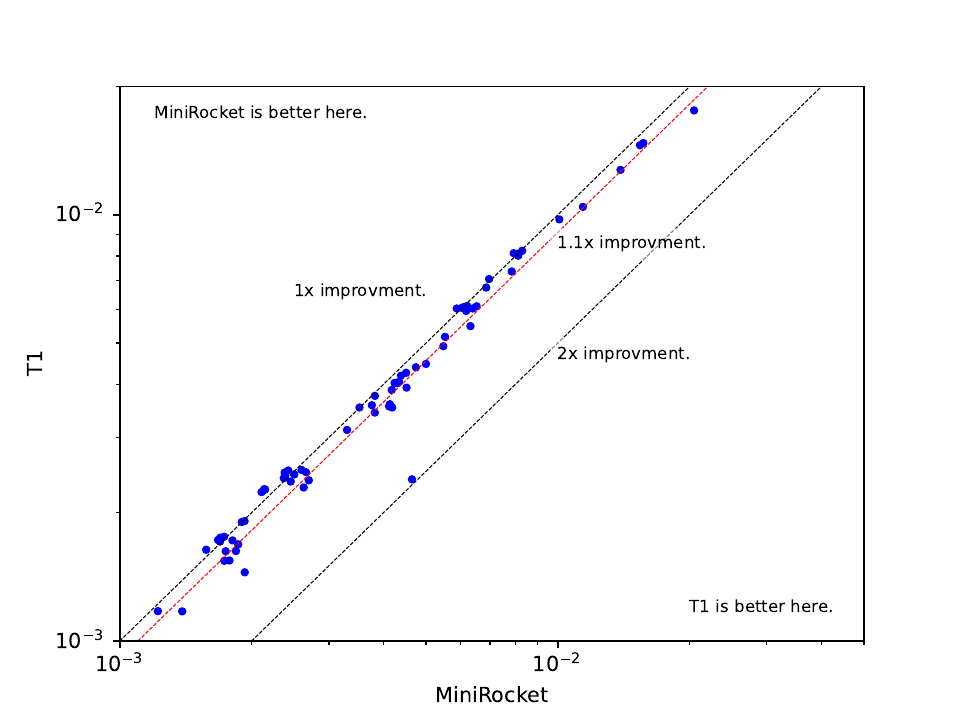}}%
 		\label{fig:t1_mini_latency}%
 	}
	\caption{Two ATCN models were pairwise compared against MiniRocket for Accuracy/Latency over 70 benchmarks of UCR 2018.}
	\label{fig:winner}
\end{figure*}

Temporal Convolutional Networks (TCN) were first proposed based on an adaptation of WaveNet \cite{WaveNet} and Time-Delay Neural Network \cite{timeDelay}. It orchestrates dilated convolutions in Encoder-Decoder architecture to have a unified framework for action segmentation. Later, Bai et al. \cite{DBLP:empirical} designed a Generic TCN (GTCN) architecture for sequence modeling, which outperforms LSTM on time-series and sequence modeling tasks. However, the GTCN suffers from two main drawbacks: 1) the size of dilation increases exponentially by the layer, which prevents the designer from increasing the depth of the network, 2) it uses two standard convolutions per each layer, which is computationally expensive for resource-constrained embedded systems. Overall, there is a lack of high-accurate lightweight deep learning solution that can be used for robust processing of time series on embedded edge devices for a broad range of embedded applications.

This article proposes an Agile Temporal Convolutional Network (ATCN) which is a novel algorithmic solution for real-time deep learning processing of time series on embedded and edge devices \footnote{The source code of the ATCN model will be publicly available at \href{https://github.com/TeCSAR-UNCC/ATCN}{https://github.com/TeCSAR-UNCC/ATCN}.}. ATCN presents a family of fast network architectures with various scaling knobs to meet a variety of design constraints. In ATCN, mirrored Spectral-Temporal Convolution Blocks (STCB) are chained in the form of Encoder-Decoder segments. In each block, residual connections are utilized to deepen the network, while separable depthwise convolutions are used to reduce the computational complexity. Fig. \ref{fig:winner} illustrates the pairwise accuracy-latency of T0 (Fig. \ref{fig:t0_mini_acc}-\ref{fig:t0_mini_latency}) and T1 (Fig. \ref{fig:t1_mini_acc}-\ref{fig:t1_mini_latency}) versus MiniRocket \cite{minirocket} which is the State-of-the-Art (SotA) network on 70 benchmarks from the UCR 2018 dataset \cite{UCRArchive2018}. The latency as explained in detail in Section \ref{sec:ER} is reported based on running benchmarks on Cortex-A57. While T0 is 1.4$\times$ faster than MiniRocket on average, it has better or equivalent accuracy for 30 benchmarks. In terms of accuracy, T1 and MiniRocket do not differ significantly from each other, but T1 does improve the latency for an average of 6\%. However, MiniRocket cannot be run on the Cortex-M7 microcontroller, T0 and T1 can be compiled and executed natively on bare-metal M7. Our experimental results also indicate that the T0 configuration reduces the MACs and model size by 102.38$\times$ and 16.84$\times$ over InceptionTime \cite{IsmailFawaz2020}, respectively. T1 also has a 73.59$\times$ reduction in MACs, and a 14.23$\times$ reduction in model size over InceptionTime.

Overall, the key contributions of this article are:

\begin{itemize}
    \item Proposing T0 and T1 as two members of the ATCN family network whose computation complexity and model size are extremely low. Additionally, the T1 model accuracy difference is statistically insignificant when compared to those of SotA networks.
    
    \item Automating the creation and training of different configurations of ATCN based on the complexity of the problem in terms of accuracy.

    \item Demonstrating the significant benefits of ATCN over SotA networks when it comes to execution on embedded IoT microcontrollers and microprocessors (ARM Cortex-M7, and Cortex-A57). 

\end{itemize}

The rest of this article is structured as follows: Section \ref{sec:RW} presents the literature review, while Section \ref{sec:BG} provides background on generic TCN and its architecture. In Section \ref{sec:atcn}, we elaborate on the Temporal-Spectral block, the ATCN architecture, and its hyperparameters. Section \ref{sec:ER} presents the experimental results including comparison with existing approaches, and finally Section \ref{sec:conl} concludes this article.

%% file: Tex/RelatedWork.tex
\section{Related Works} \label{sec:RW}
Traditional convolutional neural networks have been primarily used in computer vision applications due to their success in capturing spatial features within a two-dimensional frame. Recently, research has shown that specialized CNNs can recognize patterns in data history to predict future observations. This gives researchers interested in time-series forecasting options to choose from over RNNs, which have been regarded in the community as the established DNN for time-series predictions. In one such case, Dilated Convolutions (DC) have been shown to achieve state-of-the-art accuracy in sequence tasks. In the first use of DC, WaveNet \cite{WaveNet} was designed to synthesize raw audio waveform, and it outperforms the LSTM. Later, Lea et al. \cite{ActionDetection} proposed TCN, a unified network based on WaveNet DC, for video-based action segmentation. In the same trend, the gated DC was used for the sequence to sequence learning \cite{DBLP:convseq2seq}. The proposed approach beats deep LSTM in both execution time and accuracy. 

GTCN \cite{DBLP:empirical} is a generic architecture designed for sequence modeling. The design of GTCN was based on two main principles: 1) there shouldn't be any information leakage from future to past, 2) the network should be able to receive any arbitrary input length similar to RNN. Since the main fundamental component of GTCN is based on variable-length DC, it brought higher parallelization and flexible receptive field in comparison to RNN. Also, since the gradient flow of GTCN is different from the temporal path of RNN, it is more resistant to the problem of gradient instability. Recent approaches have taken advantage of GTCN benefits or similar architectures in their works. In the work of \cite{depthWise_Audio}, a modified version of GTCN with depth-wise convolution has been used to enhance the speech in the time domain. The DeepGLO \cite{sen2019think} is another work that used a global matrix factorization model regularized by a TCN to find global and local temporal in high dimensional time series.

InceptionTime \cite{IsmailFawaz2020} is an ensemble of CNN blocks called Inception Module and proposed as a solution for the Time Series Classification (TSC) problem. The network architecture was constructed based on the Inception-v4 \cite{Inception} structure, and it employed a larger kernel size to beat enormous and complex models such as the HIVE-COTE \cite{HIVE_COTE}. Another research article based on deep learning methodology called Elastic Matching-CNN (EM-CNN) \cite{EM_CNN} aims at enhancing famous vision CNNs, such as Inception, and ResNet, with matching convolution (MConv) to outperform SotA solutions such as HIVE-COTE for time series classification. Nonetheless, they are still exceedingly heavy for the microcontroller with limited resources, such as ARM Cortex-M series, or even embedded ARM Cortex-A series microprocessors.

Both HIVE-COTE and Rocket \cite{ROCKET} are current non-deep learning methods that are regarded as SotA for time series classification. Rocket consists of a linear classifier and signal transformers that map inputs into features by convolving them through various and random kernels. To classify the inputs, the extracted features are passed through the linear layer. MiniRocket is a variant of Rocket that minimizes training time while ensuring accuracy by reducing exploration space and using kernels with almost deterministic values to minimize transformer complexity. However, due to the non-deep learning characteristics of their models, they still suffer from the lack of standard tools and libraries to map and execute on bare-metal embedded microcontrollers.

On top of algorithmic design, pruning \cite{He_2017_ICCV, han2015deep_compression, DBLP:journals/corr/MolchanovTKAK16}, quantization \cite{Jacob_2018_CVPR, han2015deep_compression, deepdive}, and knowledge distillation \cite{Hinton2015DistillingTK} are orthogonal compression methods to enable implementation and optimization of deep learning algorithms on tightly resources constrained edge devices. McFly \cite{Mcfly} is another independent research line, which is a brute force algorithm that finds hyperparameters for existing models such as InceptionTime and ResNet to explore design and training knobs to find optimal network configuration. This paper proposes ATCN for embedded and resource-constrained hardware to address time-series domain problems, which is on par with InceptionTime in terms of accuracy performance. The unique Encoder-Decoder structure of ATCN with the use of depthwise convolution and residual connection helps the model have a better or similar performance in respect to the InceptionTime without putting a burden on final hardware. We have put ATCN on test in Section \ref{sec:ER} by comparing ATCN against InceptionTime over 70 benchmarks from UCR time-series datasets. Additionally, we have reported the execution profile of the Cortex-M7 and Cortex-A57 when running ATCN families. We have shown that ATCN improves or maintains the overall system accuracy for these three cases while minimizing computational complexity and model size. In the next section, we study the structure of DC in-depth to prepare the ground for introducing ATCN in Section \ref{sec:atcn}.

%% file: Tex/Background.tex
\section{Background: Temporal Neural Networks}\label{sec:BG}
GTCNs are designed around two basic principles: 1) the convolutional operations are causal, i.e., predictions are made based only on current and past information; 2) the network receives an input sequence of arbitrary length and maps it to an output sequence of the same length \cite{DBLP:empirical}. Based on principle number 2, in order to map the final output to an arbitrary size, the output of the last DC output can be connected to a linear layer. This adds flexibility by allowing a final output length to be independent of the input length. The naive causal convolutions, which have a dilation rate of 1, are inherently inefficient as their sequence history scales with a size linear to the depth of the network. 

\begin{figure}[!h]
	\centering
	\includegraphics[width=.55\linewidth,trim= 20 15 20 20,clip]{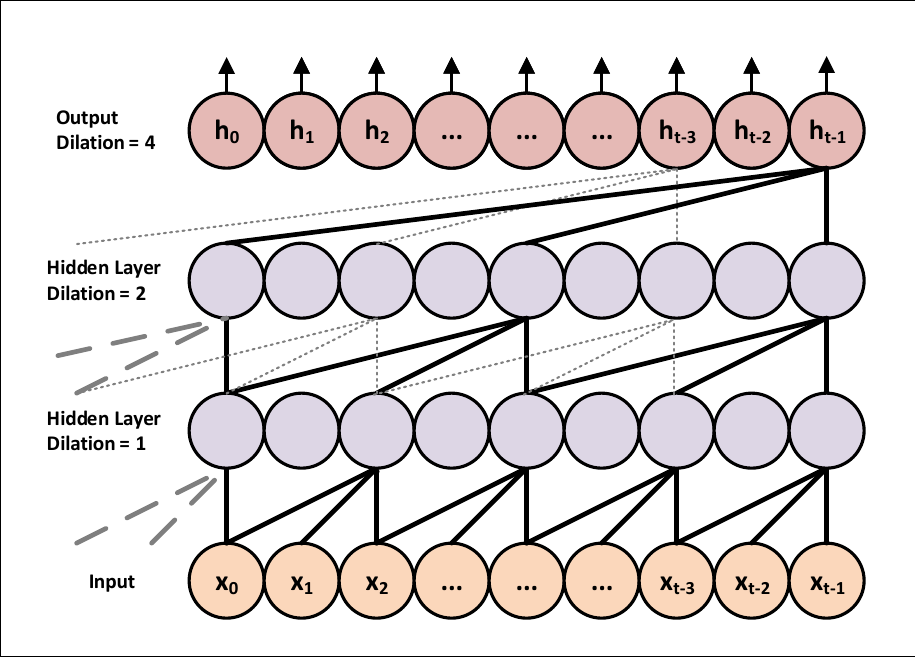}
    \caption{Dilated Causal Convolution.}
    \label{fig:dilated}
\end{figure}

\begin{figure}[!h]
	\centering
			\includegraphics[width=.45\linewidth,trim= 10 10 10 10,clip]{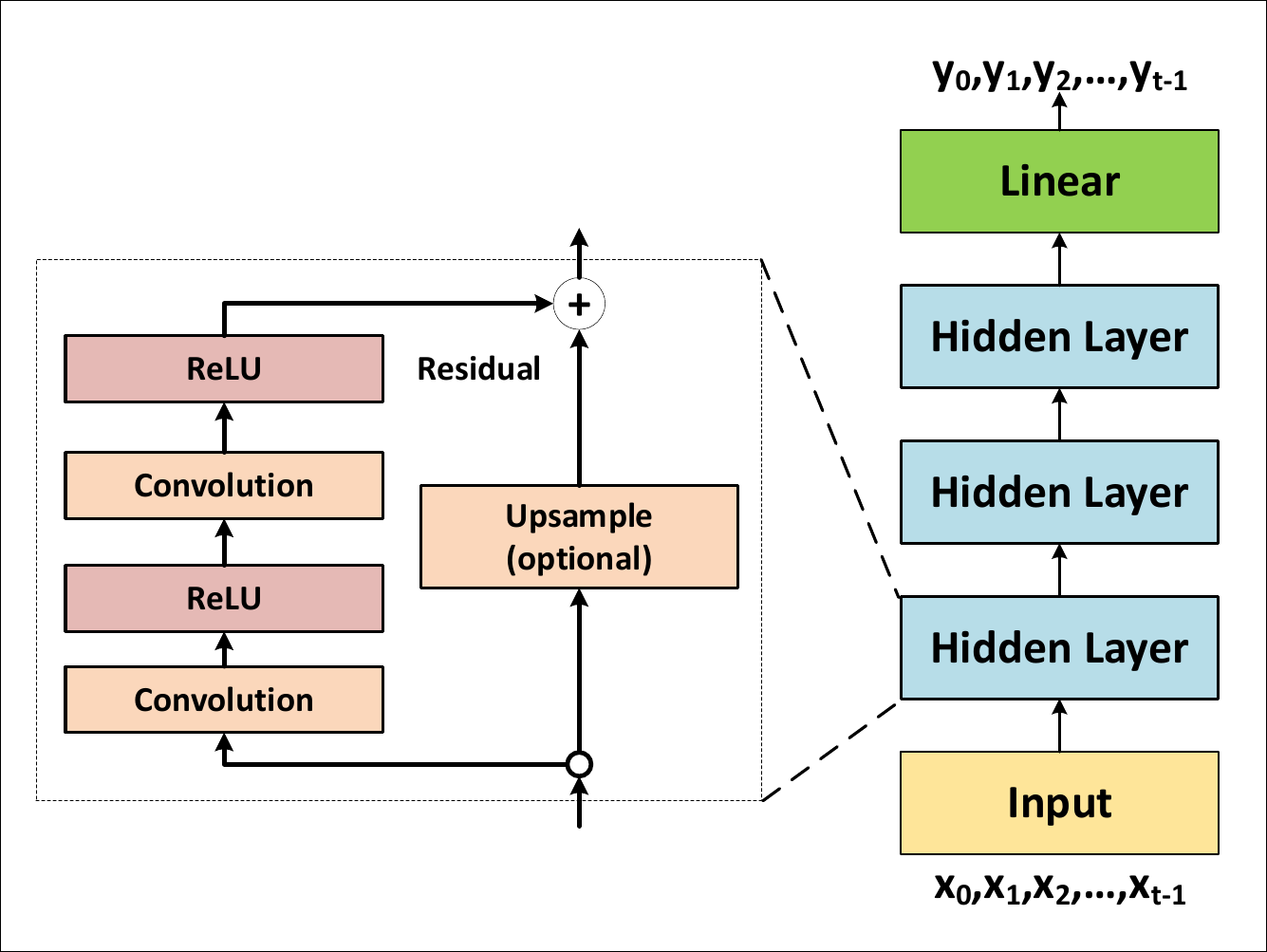}
\caption{Structure of Generic TCN.}
\label{fig:gtcn}
\end{figure}

The solution here incorporates dilated convolutions to exponentially scale the receptive field, as shown in Fig.~\ref{fig:dilated}. The kernel of a dilated convolution is stretched to cover a larger area of the input. In order to accomplish this, holes (zeroes) are placed between the kernel elements. The dilation rate defines the number of spaces that will be inserted between kernel elements, which determines the level of enlargement. The dilation rate of $d$ is generally expressed in $d-1$ spaces. Fig.~\ref{fig:dilated} shows dilated convolutions applied on input vector $X$ = $[x_0,~x_1,~x_2,...,~x_{t-1}]$. The dilation rate for layer 0 is 1, equivalent to a standard convolution. When the depth increases, the dilation rate increases to 2 and then to 4, as 1 and 3 holes are inserted in the respective kernels. The first convolution with dilation rate $d$=1  maps the input vector $X$ to the higher dimension.  Then, GTCN increases the $d$ for the next convolutions exponentially to increase the receptive field. The minimum output sequence length, before mapping to the linear layer, can be determined by calculating its receptive field: \cite{tcnfpga}:
\begin{equation}
\label{eq:receptiveField}
rf = 1 + \sum_{l=1}^{L}[k(l)-1]\times d(l),
\end{equation}
where $l \in \{1,2,3,...,L\}$ is the layers, $k$ is the kernel size, and $d(l)$ is the dilation rate at layer $l$. This means that as the depth of the network increase, so does the receptive field. The dilated convolution of $F$ on element $s$ of a sequence $X$ is given as: 
\begin{gather}
\label{eq:dilated}
F(s) = (X*_{d}f)(s) = \sum_{i=0}^{k-1}f(i)\cdot x_{s-d \cdot i},
\end{gather}
where $X \in \textbf{R}^{n}$ is a 1-D input sequence, $*_{d}$ is dilated convolution operator, $f : \{0, ..., k-1\} \in \textbf{R}$ is a kernel of size $k$ and $d$ is the dilation rate \cite{DBLP:empirical, tcnfpga}. For applications requiring a very large $rf$, it is also essential to provide stability in the later layers subject to the vanishing gradient problem. A popular technique in traditional CNN architectures, the residual block \cite{resnet}, provides a “highway” free of any gated functions, allowing information to flow from the early layers to the last layers unhindered. These connections can be seen in the final GTCN architecture shown in Fig.~\ref{fig:gtcn}. The GTCN consists of $L$ hidden layer and an optional linear layer to map the input size $i$ to arbitrary output size. In order to increase the capacity of the network to absorb more features from its input, each hidden layer of GTCN has two regular convolutions and two ReLU activation functions. There can also be an upsampling unit, such as point-wise convolution, in the first hidden layer of GTCN to map 1-D input sequence to a higher dimension to guarantee the element-wise addition receives tensor of the same dimension.

The design of GTCN suffers from two problems: 1) exponential growth of dilation size, 2) the existence of two regular convolutions per layer. The exponential growth of dilation size and requirement of having the same length for both input and output of dilated convolution force the network designers to have excessive padding at the higher layers. Also, the implementation of two convolutions blocks per layer makes the GTCN very expensive computationally. In the next section, we address the problems mentioned above by introducing ATCN architecture.

%% file: Tex/Architecture.tex
\section{ATCN: Agile Temporal Convolutional Networks}\label{sec:atcn}
In this section, we introduce the architecture of ATCN. At first, we discuss the essential components, and then we elaborate on the hyper-parameters, and in the end, we present the ATCN architecture and its model builder.

\subsection{Network Structure}  
\begin{figure}[t]
	\centering
	
	\subfigure[Group convolution in STCB is set to depthwise convolution]{\includegraphics[width=.33\linewidth,trim= 15 15 20 15,clip]{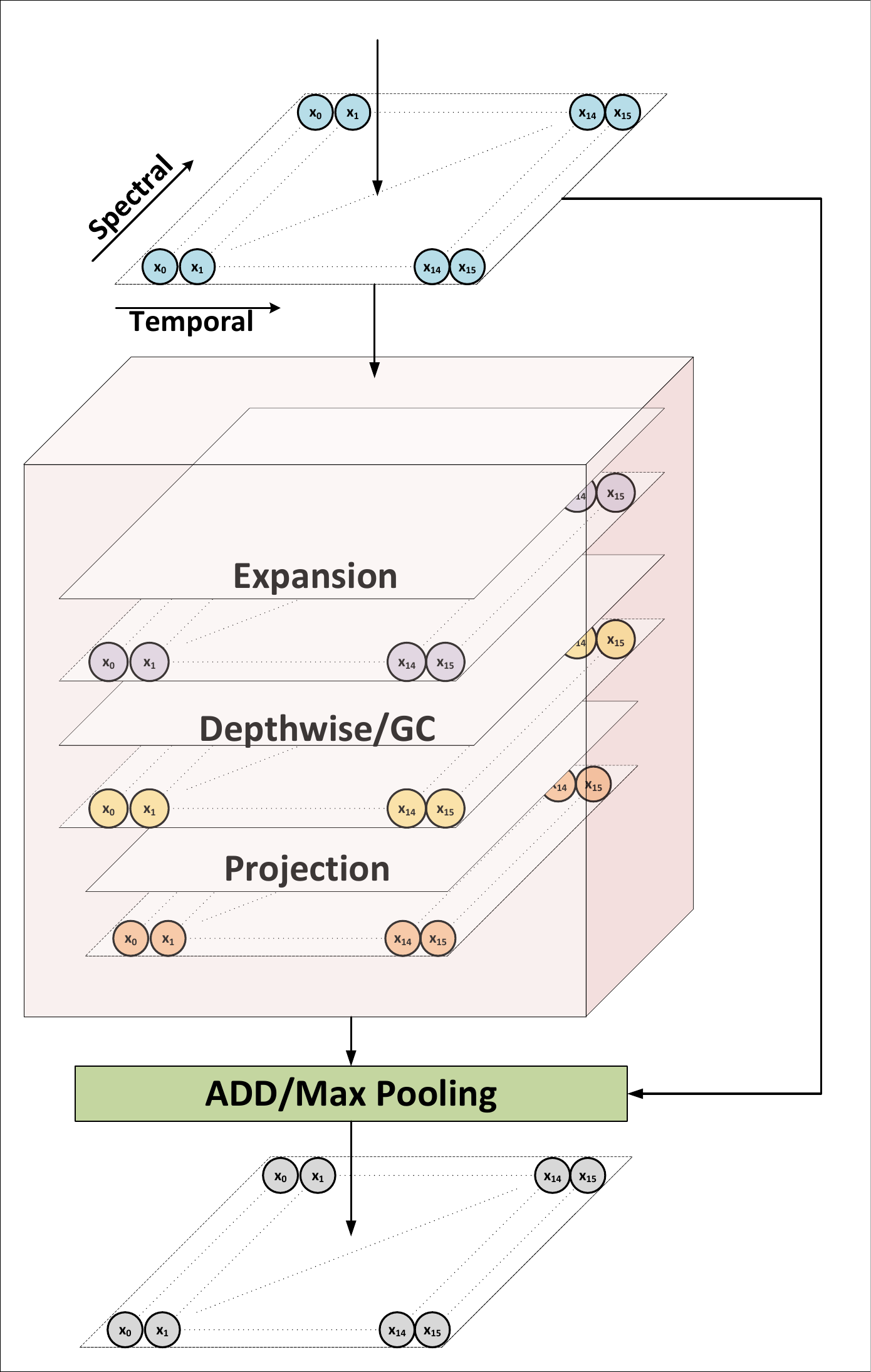}\label{fig:atcnblocks_dw}}\quad \hspace{1.cm}
	\subfigure[ATCN Architecture]{\includegraphics[width=.235\linewidth,trim= 20 15 20 15,clip]{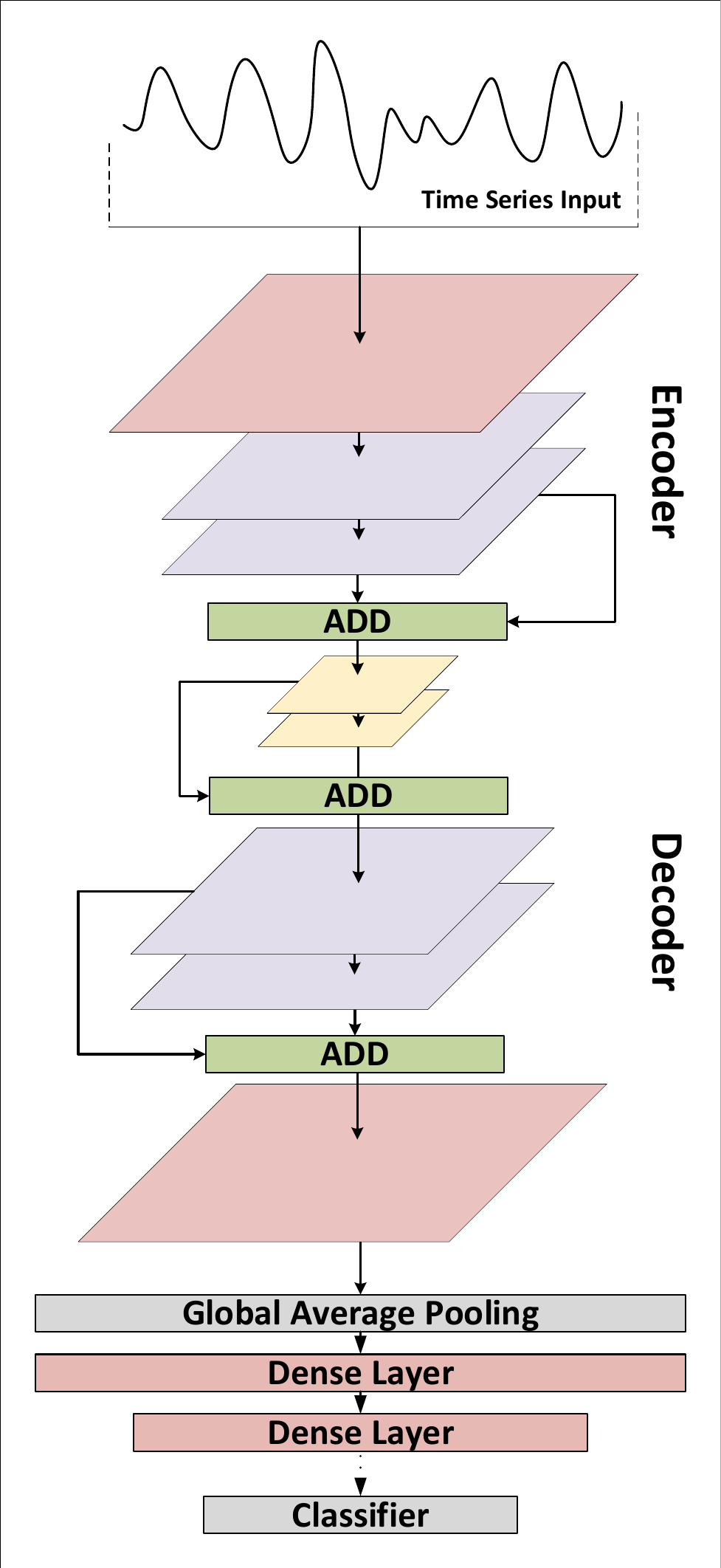}
	\label{fig:atcnblocks_arch}
	}
	\caption{Structure of STCN blocks (\ref{fig:atcnblocks_dw}) and ATCN architecture (\ref{fig:atcnblocks_arch}). The non-linearity activation and batch normalization units after each convolution are not depicted. The structure of STCB with a group (depthwise) convolution and MaxPooling (Add) unit is depicted in \ref{fig:atcnblocks_dw}. For the STCB with the depthwise convolution, a skip-line is added, since the input size and residual path have the same dimension. In STCB configuration, the purple and orange output is the result of pointwise convolution, and the yellow output is the result of group (depthwise) convolution.}
	\label{fig:atcnblocks}
	\vspace{-10pt}
\end{figure}

At the core, the architecture of ATCN is a chain of Spectral-Temporal Convolution Blocks (STCB). Each STCB is composed of a pointwise (expansion), a group, and a pointwise (projection) convolution. We visualized the STCB in Fig.~\ref{fig:atcnblocks_dw}. The MaxPooling layers are optional. It lets architects downsample temporal information to minimize computational complexity while embedding that information in higher or lower dimensions. The extreme case of STCB is when the group size and its input channel size are equal. In this case, the group-convolution is set to depthwise, and the ATCN network synthesizer will remove the maximum pooling and add a skip-line between element-wise addition and the STCB input. The final architecture of ATCN is shown in Fig. \ref{fig:atcnblocks_arch}. The ATCN is a mirrored residual dilated convolutional neural network. It starts with mapping the $n$-dimensional input, which is generally 1-D for the time series, to a higher dimension at the first layer. Then, it encodes the data to lower dimensions. At decoder part, the data will be decoded to a higher dimension again. Based on the final application, the output of the decoder can be used for regression or classification problems. In the rest, we discuss the details of STCB and the network hyper-parameters.

\begin{figure}[!h]
	\centering
	\subfigure[Effect of different non-linearity activation function on validation loss.]{
		\includegraphics[width=.45\linewidth,trim= 1 1 0 1,clip]{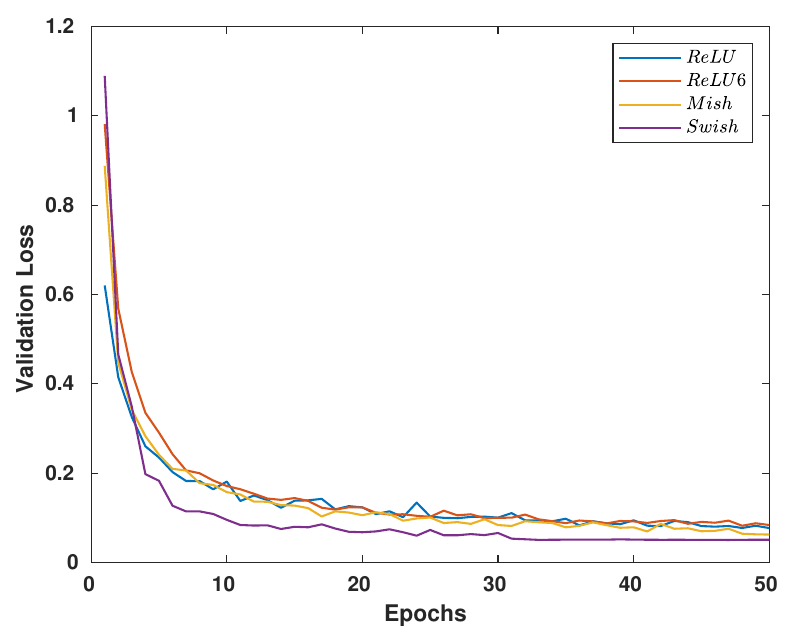}
	\label{fig:act_effect}
	}
	\subfigure[Dilated Causal Convolution.]{	
	\includegraphics[width=.45\linewidth,trim= 1 1 0 1,clip]{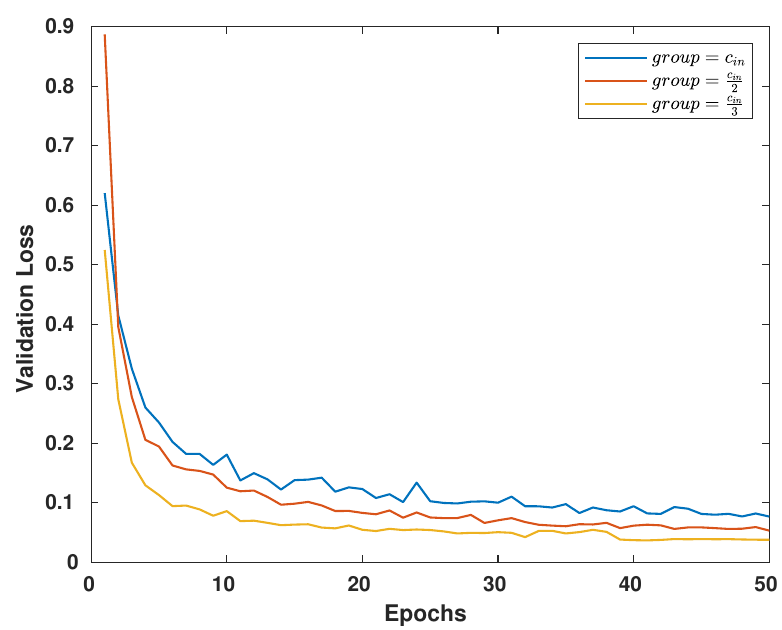}
	\label{fig:group_effect}
	}
\caption{The effect of different non-linearity activation function, $\sigma$, and $group$ value on final validation loss.}
\end{figure}

The first layer of ATCN is standard convolution with an optional MaxPooling for downsampling the input. A padding unit is also added before standard convolution and expansion convolution in STCB to ensure that the input and output tensors of the block have the same size to satisfy principle number 2 of GTCN. The $2p$ zeros are added symmetrically by padding unit, where $p$ is given by:
\begin{gather}
    p = \lceil \frac{(o-1)\times s+(k-1)\times(d-1) -i+k }{2} \rceil,
    \label{eq:padding}
\end{gather}
where $o$ is the output size, $i$ is the input size, $s$ is the stride, $k$ is the kernel, and $d$ is the dilation. After each convolution, 1D batch normalization and a non-linear activation function is also added. In this paper, we used $Swish$ as activation function:
\begin{gather}
    Swish(X) = X \odot Sigmoid(X),
\end{gather}
where $\odot$ is Hadamard or element-wise multiplication. Fig. \ref{fig:act_effect} shows how different activation functions perform on the validation loss of the MNIST digits classification problem. It should be noted that based on the complexity of the application and available memory and hardware computational power, the activation function of ATCN can be changed.

The STCB consists of expansion, followed by a group and another projection convolution. The task of expansion convolution is to map input channel size, $c_{in}$, to higher or same dimension, $c^{exp}_{out}$, where $c^{exp}_{out}=\alpha \times c_{in}, \alpha \geq 1$. On the contrary, pointwise projection embeds and maps the feature extracted from the group convolution to the block output size, $c_{out}$. For the case of depthwise convolution, we set $group$, which manages the connection between input and output, to $c^{exp}_{out}$. For this case, the convolution weigh shape changes from $(c_{out}, c_{in}, k)$ to $(c_{out}, 1, k)$, where $k$ is the kernel size. We designed the network synthesizer so that if $c_{in} = c_{out}$, the skip line is automatically created from input to the elementwise addition. Then, the input will be added to the residual output from the pointwise projection. The residual connection helps the designers increase the network's depth without being worried about the vanishing gradient problem. The model synthesizer considers group convolution rather than depthwise for the case that MaxPooling is selected. The reason for doing so is based on this observation that for downsampling the input, which has an activated max-pooling unit, group convolution helps to better map temporal information to a higher dimension without drastically increasing computation complexity and the model size. The only constraint imposed by group convolution is that its output channel size, $c^{gc}_{out}$, should be divisible by $c^{exp}_{out}$. The two extreme G-CNN cases are when $group=c^{gc}_{in}$ and $group=1$. In the former case, the group convolution is a depthwise convolution, and in the latter, it is a standard convolution. Formally, the weight shape for group convolution is $(c_{out}, \frac{c_{in}}{group}, k)$. We depict the effect of altering the $group$ values in Fig.~\ref{fig:group_effect} for MNIST digit classification. As we can see, reducing the $group$ value increases the network capacity to minimize the validation loss.

\subsection{The Receptive Field of an ATCN} \label{subsec:HP}
To maintain the receptive field of an ATCN network at different network depths, two knobs must be adjusted: 1) kernel size, and 2) dilation rate. We elaborate on each of them in detail in the rest of the article.

\subsubsection{Dilation rate} For a fixed input size, if we increase the number of layers, based on the GTCN architecture guideline, we need to increase the dilation rate exponentially. This decision will help the network have a higher receptive field; however, based on principle number 2, we need to pad the features excessively to have the same input and output size. This unnecessary padding results in 1) more computation and 2) CNN performance degradation. We observed linear growth for dilation would help the network with more than six layers to have better feature representation. Although the dilation rate can be defined as a function of layer number, we increased it after each block with activated downsampling in the experimental results. This decision helps design a deep ATCN for the cases where input size, $i$, is small.

\subsubsection{Kernel size}\label{subsub:ks} It is recommended that the kernel size, $k$, is large enough to encompass enough feature context based on the problem complexity. However, based on Eq. ~\ref{eq:padding}, it is a good practice to decrease the kernel size for higher layers to ensure $p$ is not growing exponentially. By contrast, if we increase the dilation rate, based on Eq. \ref{eq:receptiveField}, the kernel size can be reduced without concern for the receptive field. Embedded devices gain two crucial advantages from this decision: it reduces 1) the computational complexity and 2) the model size.

Similar to the dilation rate, if we need to increase the network's depth to enhance its capacity, it is recommended to gradually decrease kernel size and have a linear growth for dilation. We can alter both after each block with downsampling units. This decision helps the final structure to have enough receptive field to cover feature context without increasing the MAC operations and model sizes.

\subsection{ATCN Model Synthesizer}
\begin{figure}[b]
	\centering
		\includegraphics[width=.55\linewidth,trim= 10 10 10 10,clip]{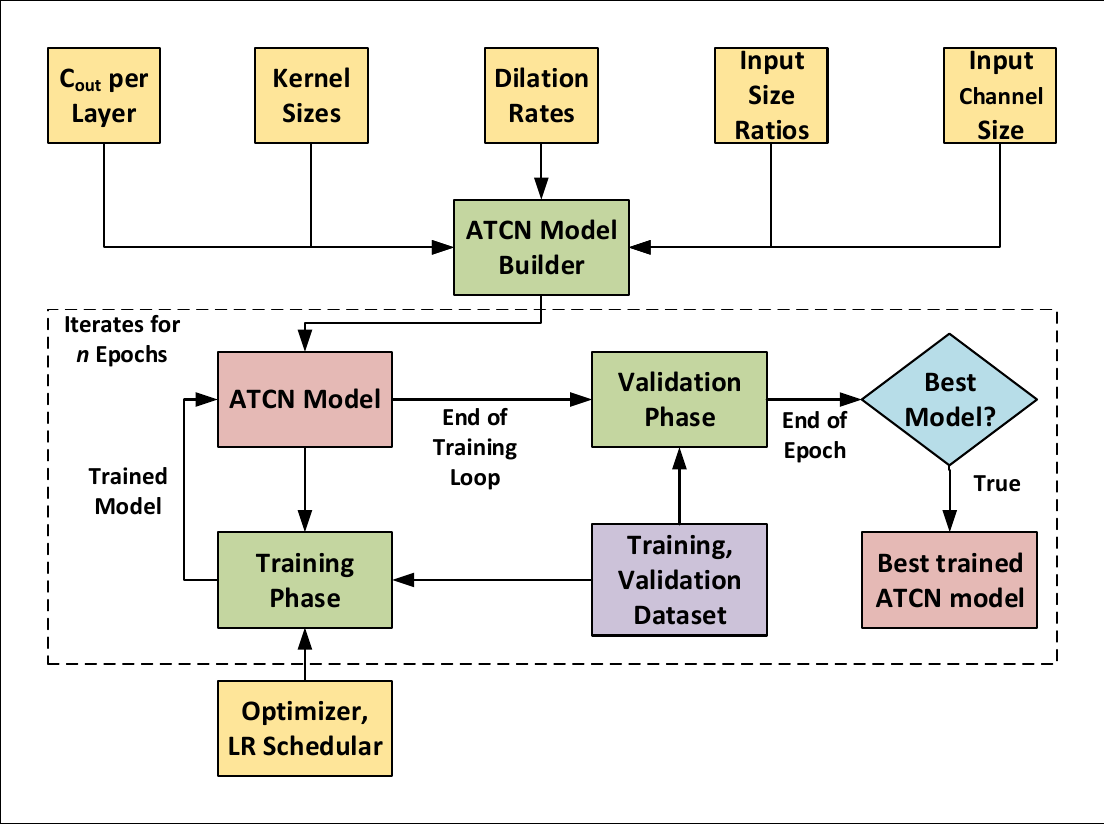}
	\caption{ATCN model synthesizer and training framework.}
	\label{fig:atcn_builder}
\end{figure}

We depict the inputs of \textit{ATCN Model Synthesizer} and the framework for its training in Fig.~\ref{fig:atcn_builder}. The ATCN Model Synthesizer receives \textit{Output Channels}, \textit{Kernel Sizes}, \textit{Dilation Rates}, \textit{Input Ratios}, and finally, the first \textit{Input Channel Size} to design the ATCN network architecture.  The model synthesizer automatically configures the STCBs and sequences them to generate the final network model based on the ATCN configurations, such as kernel size, and dilation rate. The model will be trained based on the provided dataset and training hyperparameters, such as learning rate and weight decay. Compression algorithms such as online quantization and pruning are orthogonal to ATCN models. Therefore, the training phase of the synthesizer can be expanded to compress the model online or offline after the training phase. The trained ATCN model can be exported as Open Neural Network Exchange (ONNX) so it can be deployed on embedded edge devices. The trained model cannot be reconfigured during the run-time as the training phase is required again to fine-tune the revised model. Ultimately, the agility of ATCN comes from the model synthesizer, which can quickly generate the ATCN model and configure the training framework to shrink design time. In the rest, we explain each of the ATCN Model Synthesizer inputs in detail.

\subsubsection{Output Channels}: It is a vector of size $L$, where $L$ is the number of layers (blocks). The $c^{l}_{out}$ defined in $\boldsymbol{C_{out}}=[c^{1}_{out}, c^{2}_{out}, ..., c^{L}_{out}]$, decides the output channel for layer $l$. For this research, the values in vector $C_{out}$ are descending-ascending (Auto-Encoder architecture) to code the feature to lower dimension to extract temporal correlation and then maps to a higher dimension to represent extracted features for final stages.

\subsubsection{Kernel Size}: The vector $\boldsymbol{K}$, $\boldsymbol{K}=[k^{1}, k^{2}, ..., k^{L}]~|~k^{l} \in \mathbb{N}$, defines the kernel sizes for each layer. Based on the discussion of Section \ref{subsub:ks}, it is suited to decrease the $k$ to minimize both model size and required computational complexity.

\subsubsection{Dilation Rates}: The vector $\boldsymbol{D}$, where $\boldsymbol{D}=[d^{1}, d^{2}, ..., d^{L}]~|~d^{l} \in \mathbb{N}$, defines the dilation rates per each layer. On the contrary to $K$, it is necessary to increase $d$ to achieve a higher or same receptive field at deeper levels.

\subsubsection{Input Size Ratios}: The $\boldsymbol{R}=[r^{1}, r^{2}, ..., r^{L}]~|~0<r^{l}\leq1$, defines the input ratios. For the value of $r^{l} < 1, l > 1$, the \textit{ATCN Model Synthesizer} configures the STCB block with max-pooling unit. For the case of $l=1, r<1$, the max-pooling will be added after standard convolution; otherwise, the input and out of standard convolution will have the same size. For this research, the $r^l$ can only be defined as $\frac{1}{2}$ or $1$. For other ratios, the synthesizer can be modified to change the stride of max-pooling to satisfy the targeted ratio.
\subsection{ATCN Families}
We introduce two ATCN families by assigning values to the $\boldsymbol{C}$, $\boldsymbol{K}$, $\boldsymbol{D}$, and $\boldsymbol{R}$ for UCR time series classification. Table \ref{tab:atcn_fam} summarizes the configuration of T0 and T1 as two candidates. We assign the output channels, $C_{out}$, in descending-ascending format to encode features to lower dimensions, then decode them to higher dimensions to represent them for the final dense layers and classifier. We have reduced the kernel size $K$ due to the increased dilation rate, $D$. As a result, MACs and model parameters are reduced without compromising receptive field size. The values assigned in Table \ref{tab:atcn_fam} per each hyperparameter of the $\boldsymbol{C}$, $\boldsymbol{K}$, $\boldsymbol{D}$, and $\boldsymbol{R}$ are considered based on input characteristics of UCR 2018 benchmarks and the resources available on targeted hardware platforms. Therefore, they can be fitted in hardware memory and executed based on available computation power while having a comparable accuracy in respect to SotA time series classifiers. Based on the hyperparameter trade-off explained in Section \ref{subsec:HP}, the designer can explicitly increase the network capacity and define the ATCN models as server-class solutions. Table \ref{tab:atcn_flops} compares the average FLOPS and number of model parameters of candidates and InceptionTime based on seventy benchmarks from the 2018 UCR time series classification, which are explained in detail in Section \ref{sec:ER}. We were not able to calculate the MACs and FLOPS of MiniRocket with the aid of standard libraries due to the non-deep learning characteristics of MiniRocket. The T0 configuration reduces the MACs and model size by 102.38$\times$ and 16.84$\times$ over IT, respectively. T1 has also 73.59$\times$ reduction in MACs, and a 14.23$\times$ reduction in model size over IT. The algorithmic accuracy of these models and training methods of these models are explained in Section \ref{sec:ER}.
\input{Tables/atcn_fam}
\input{Tables/atcn_flops}

%% file: Tables/atcn_fam.tex
\begin{table}[htbp]
  \centering
  \caption{The configuration of two ATCN families}
    \begin{tabular}{cllll}
    \toprule
    \multicolumn{1}{c}{\multirow{2}[4]{*}{\textbf{Models}}} & \multicolumn{4}{c}{\textbf{Configurations}} \\
\cmidrule{2-5}          & \multicolumn{1}{c}{$\boldsymbol{C_{out}}$} & \multicolumn{1}{c}{$\boldsymbol{D}$} & \multicolumn{1}{c}{$\boldsymbol{K}$} & \multicolumn{1}{c}{$\boldsymbol{R}$} \\
    \midrule
    T0    & $[32, 16, 16, 8, 8, 16, 16, 32]$ & $[1, 2, 2, 4, 4, 6, 6, 8]$ & $[32, 16, 16, 8, 8, 4, 4, 2]$ & $[\frac{1}{2},  1,   1,  1,  1,  1,  1, 1]$ \\
    T1    & $[32, 16, 16, 8, 8, 16, 16, 32]$ & $[1, 2,  2, 4, 4, 6, 6, 8]$ & $[64, 32, 32, 16, 16, 8, 8, 4]$ & $[\frac{1}{2},  1,   1,  1,  1,  1,  1, 1]$ \\
    \bottomrule
    \end{tabular}%
  \label{tab:atcn_fam}%
\end{table}%

%% file: Tables/atcn_flops.tex
\begin{table}[htbp]
  \centering
  \caption{FLOPS and number of parameters for T0, T1, and InceptionTime.}
    \begin{tabular}{lccc}
    \toprule
    \multicolumn{1}{c}{\multirow{2}[3]{*}{\textbf{Metric}}} & \multicolumn{3}{c}{\textbf{Models}} \\
\cmidrule{2-4}          & \multicolumn{1}{c}{\textbf{T0}} & \multicolumn{1}{c}{\textbf{T1}} & \multicolumn{1}{c}{\textbf{InceptionTime}} \\
    \midrule
    \textbf{FLOPs} & 2,377,840 & 3,329,008 & 240,430,566 \\
    \textbf{Params\#} & 24,816 & 29,424 & 422,498 \\
    \bottomrule
    \end{tabular}%
  \label{tab:atcn_flops}%
\end{table}%

%% file: Tex/ExpResults.tex
\section{Experimental Results}\label{sec:ER}
In this section, we demonstrate the capabilities of two different ATCN families by applying them to problems of UCR time series classification. T0 and T1 are compiled and utilized on both ARM Cortex-M7 microcontroller and Cortex-A57 microprocessor. A report on RAM utilization, flash usage, and inference time is also provided. In following Demvsar's recommendation \cite{demvsar2006statistical}, we used an open-source tool \footnote{\href{https://github.com/hfawaz/cd-diagram}{Available at: https://github.com/hfawaz/cd-diagram}}
 to repeat the comparison setup explained in \cite{IsmailFawaz2018deep} to practice the Friedman test \cite{friedman1940comparison} and use Wilcoxon signed-rank test \cite{wilcoxon1992individual} with Holm’s alpha (5\%) correction \cite{Holm1979} to compare the accuracy and latency of each model among the benchmarks for Fig. \ref{fig:cdLatency} and Fig. \ref{fig:cdd}. We reported Top-1 accuracy as the final model accuracy and the latency is the combined delay between an input and its classification output when the batch size is one.

\subsection{Dataset}
The experiments were conducted on 70 benchmarks \footnote{A list of the benchmarks are mentioned in Appendix \ref{sec:app} and will also be available online for the sake of reproducibility.} publicly available from UCR Time Series Classification 2018, which vary in time length, number of classes, dataset type, and sample size. As our article focuses on embedded devices, we chose benchmarks based on the fact that they are typically observed on edge devices. Therefore, we selected benchmarks whose types are image, spectrum, ECG, or sensors while covering a wide range of input lengths and the number of classes.

\subsubsection{Data augmentation}
\begin{figure*}[h]
	\centering
	
	\subfigure[Jittering]{%
		{\includegraphics[width=.35\linewidth, trim= 1 1 1 1,clip, keepaspectratio]{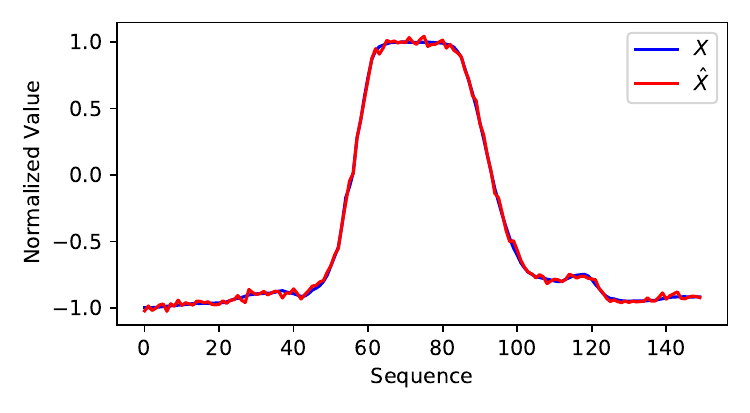}}%
		\label{fig:Jittering}%
	}\qquad
 	\subfigure[Magnitude Warping]{%
 		{\includegraphics[width=0.35\textwidth, trim= 1 1 1 1,clip, keepaspectratio]{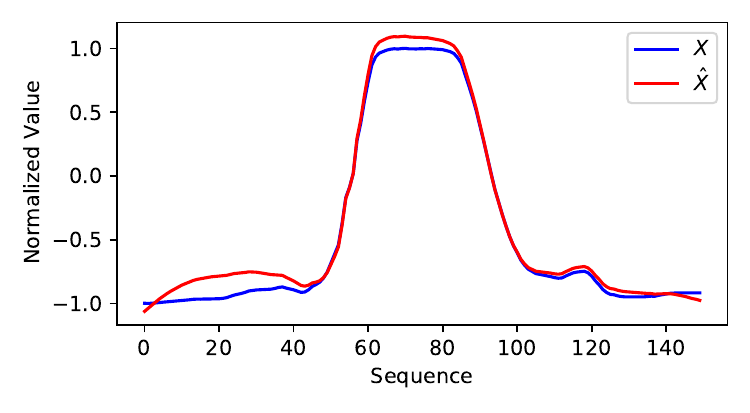}}%
 		\label{fig:mw}%
 	}
 	\\
	\subfigure[Window Wraping]{%
		{\includegraphics[width=.35\linewidth, trim= 1 1 1 1,clip, keepaspectratio]{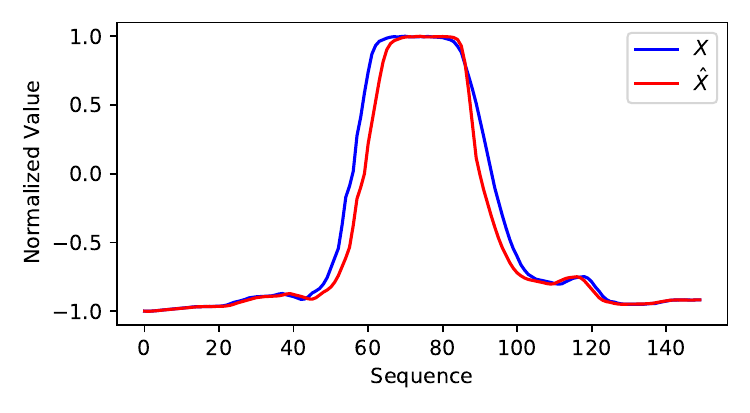}}%
		\label{fig:ww}%
	}\qquad
 	\subfigure[Scaling]{%
 		{\includegraphics[width=0.35\textwidth, trim= 1 1 1 1,clip, keepaspectratio]{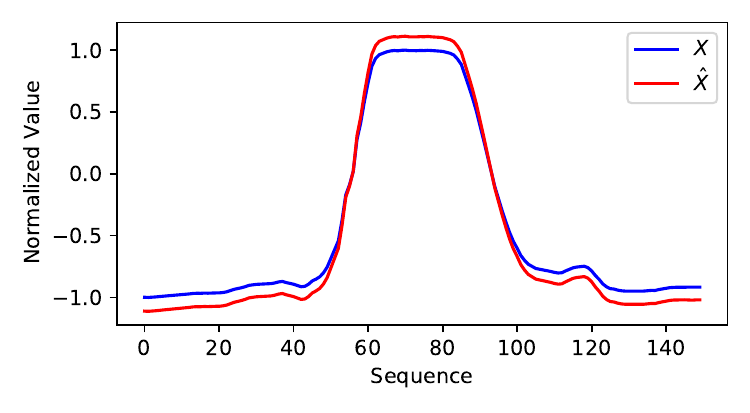}}%
 		\label{fig:sc}%
 	}
	\caption{Different data augmentation applied on UCR dataset. $X$ is observed signal and $\hat{X}$ is the augmented data.}
	\label{fig:data_aug}
\end{figure*}

For benchmarks whose training size is small, such as \textit{ECGFiveDays}, we applied four types of data augmentation: jittering \cite{Iwana_2021}, magnitude warping \cite{data_aug_mw}, window warping \cite{data_aug_ww}, and scaling. In Fig. \ref{fig:data_aug}, each approach is shown in relation to the observed signal, $X$.

\subsection{Implementation details}
The models are implemented in PyTorch and trained on an Nvidia Tesla V100 GPU using the ADAM optimizer with a Learning Rate (LR) of 0.001, a gradient clip of 0.25, and a weight decay of 0.001. We also reduce the LR by the factor of 0.1 when the validation loss stagnates for eight epochs. In the case of datasets with two classes, Binary Cross-Entropy (BCE) loss function is used, and Cross-Entropy is used for all other datasets.

\subsection{Execution Comparison}
In order to evaluate the performance of models on bare-metal embedded devices, we selected \texttt{STM32F746ZGT6}, which has Cortex-M7 running at 216 MHz with 320 KB of RAM and 1 MB of flash memory as a microcontroller. The networks have been compiled and 8-bit quantized with the aid of network compiler provided by \texttt{STM32CubeIDE}. Our goal was to run the models on M7 without any support for Real-Time Operating Systems (RTOS) and memory management. We compared the execution performance and hardware utilization of all four classifiers in Table \ref{tab:run_time}. The 8-bit quantized InceptionTime model could not be executed due to its higher RAM and flash requirements. Furthermore, we cannot use standard Cortex-M7 compiler tools to map MiniRocket to the microcontroller as it relies on Numba as a high-performance Just-In-Time (JIT) Python compiler and Thread Building Block (TBB) library. The results are extracted by a running model trained on \textit{Coffee} benchmark. Compared to the T1 configuration, the T0 can reduce both RAM and flash utilization by 7.5\% and 1.47\%, respectively, while improving inference time by 21.42\%. As a result of the low memory and computational resources requirements, ATCN models were the only ones capable of running on tiny microcontrollers.  

\input{Tables/runTime}

\begin{figure}[h]
    \centering
    \includegraphics[width=0.55\textwidth, trim= 1 1 1 1,clip, keepaspectratio]{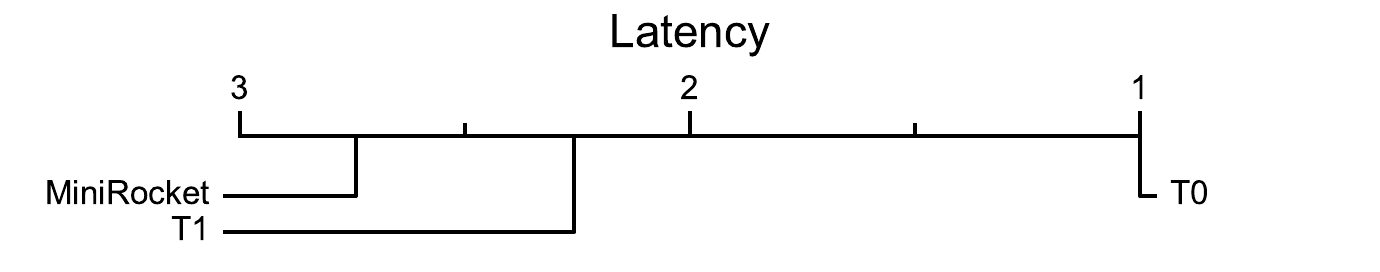}    
    \caption{Mean rank of T0 and T1 in terms of latency against MiniRocket over 100 resamplings of running seventy benchmarks from UCR 2018 on Cortex-A57. T0 outperformed the other two models and ranked first in all seventy benchmarks.}
    \label{fig:cdLatency}
\end{figure}

For Cortex-A57, we have Ubuntu 18.04 set up as the operating system with 4GB RAM and 64GB flash memory. In addition, we used \texttt{onnxruntime 1.4.0} to run ATCN ONNXes for all seventy benchmarks. The CPU is set as the execution engine for the \texttt{onnxruntime} module. To run MiniRocket on A57 processors, we compiled \texttt{LLVM 10.0.1} and \texttt{oneAPI TBB 2021.4} for the \texttt{Aarch64} architecture since MiniRocket relies on the \texttt{Numba} JIT Python compiler. We resampled the latency 100 times per benchmark and skipped the first 5 iterations as a warm-up, and the averaged latency of over 95 iterations was considered. According to Fig. \ref{fig:cdLatency}, T0 ranked first because it outperformed T1 and MiniRocket in all seventy benchmarks.


\subsection{Algorithmic Comparison}



\begin{figure}[h]
    \centering
    \includegraphics[width=0.55\textwidth, trim= 1 1 1 1,clip, keepaspectratio]{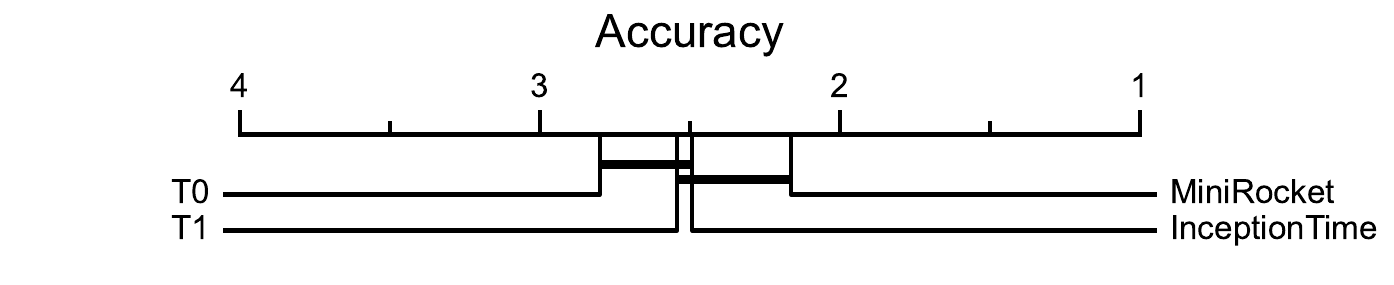}    
    \caption{Mean rank of T0, T1, InceptionTime, and MiniRocket. The diagram represents the overall average ranking of the classifiers, with a thick horizontal line indicating a group of classifiers that are not significantly different from each other in terms of accuracy. Consequently, T1, InceptionTime, and MiniRocket are not significantly different.}
    \label{fig:cdd}
\end{figure}

In Fig. \ref{fig:cdd}, we show the critical difference diagram of discussed classifiers. The connected classifiers by a thick line indicate that they do not have a significant difference statistically based on the Friedman test. In terms of accuracy, MiniRocket outperforms T0, T1, and InceptionTime; however, it shows similar performance to InceptionTime and T1 statistically. Likewise, T0, T1, and InceptionTime have similar performance. Fig. \ref{fig:cdd} also indicates that T0 and T1 have similar or better accuracy than MiniRocket for some of the benchmarks as the MiniRocket does not outperform these models in all seventy benchmarks. Since T0 and T1 have a lower latency, we need to extract the Pareto optimal designs to have efficient models in terms of latency and accuracy. 

\subsection{Pareto solutions}
\begin{figure}[t]
    \centering
    \includegraphics[width=.65\textwidth, trim= 1 1 1 1,clip, keepaspectratio]{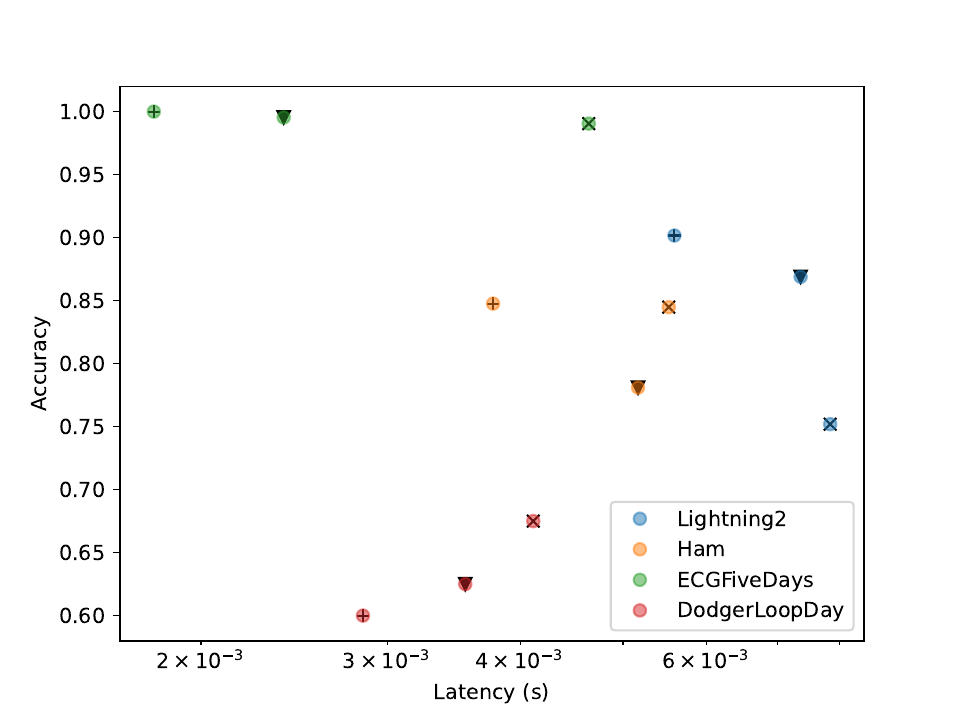}    
    \caption{The accuracy-latency Pareto front of four sample benchmarks for three models, T0 ($+$), T1 ($\blacktriangledown$), and MiniRocket ($\times$). Three examples of T0 unique solutions include \textit{Lightning2}, \textit{Ham}, and \textit{ECGFiveDays}. As can be seen, T0 produces results with higher or equal accuracy while its latency is lower than the other two models. As a result, T1 and MiniRockets do not fall under Pareto solutions. The \textit{DodgerLoopDay} benchmark is also used as an example where all three models have Pareto solutions.}
    \label{fig:pareto}
\end{figure}

In this section, we draw the accuracy-latency Pareto front based on the performance of T0, T1, and MiniRocket models. InceptionTime is ignored in this section since its accuracy is less than MiniRocket and its model complexity is higher than T0, and T1 in both FLOPS and model size metrics. To ensure each model generates meaningful results, we calculated the Pareto front with the constraint that accuracy must be at least 55\%. This constraint can be modified based on application requirements. Fig. \ref{fig:pareto} illustrates the final 4 solutions and we have not shown all 144 Pareto points for the sake of clarity. To demonstrate how the three models can provide optimal implementations, we select the \textit{DodgerLoopDay} benchmark. We can see how three models can generate three different solutions for the \textit{DodgerLoopDay} based on distinct accuracies and latencies. Table \ref{tab:pareto} summerizes the contribution of each model to provide the overall 144 optimal points. When a model provides equivalent or higher accuracy and less latency than all other models, we assign it to the \textbf{Unique} set. Those solutions provided by a model that meet the defined constraints and are part of the Pareto front make up the \textbf{Total} set of the model. On 144 Pareto solutions, T0 contributed 66 points, and it outperformed T1, and MiniRocket in 15 benchmarks. Fig. \ref{fig:pareto} shows \textit{Lightning2}, \textit{Ham}, and \textit{ECGFiveDays}, which were selected from T0 unique solutions. As we can see, for all T0 selected unique solutions, not only does T0 have better accuracy than MiniRocket and T1, but it is also faster. While T1 and MiniRocket each contribute 39 points, MiniRocket managed to beat T0 and T1 for two benchmarks named \textit{RefrigerationDevices} and \textit{EOGVerticalSignal}. Due to T0 and T1 accuracy being less than 55\% for both benchmarks, the T0 and T1 solutions were omitted.

\input{Tables/pareto}

\subsection{Architectural configuration study}
\begin{figure}[h]
    \centering
    \includegraphics[width=0.55\textwidth, trim= 1 1 1 1,clip, keepaspectratio]{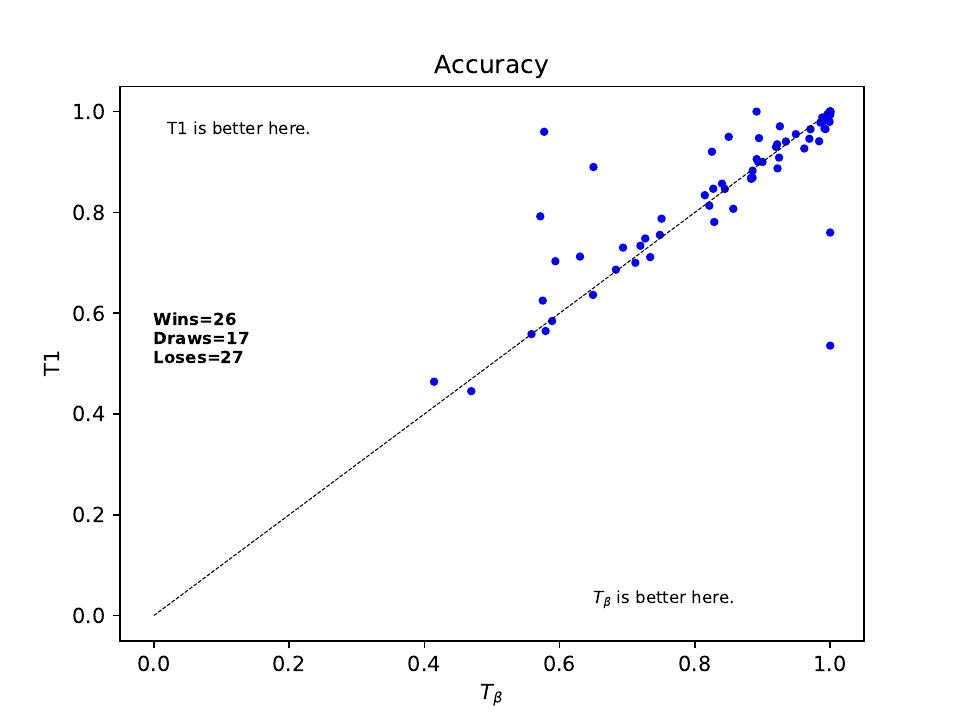}    
    \caption{Pairwise accuracy of T1 versus $T_{\beta}$.}
    \label{fig:t1_tbeta}
\end{figure}

\begin{figure*}[t]
	\centering
	
	\subfigure[Input and T0 CAM output for class 0]{%
		{\includegraphics[width=.30\linewidth, trim= 1 1 1 1,clip, keepaspectratio]{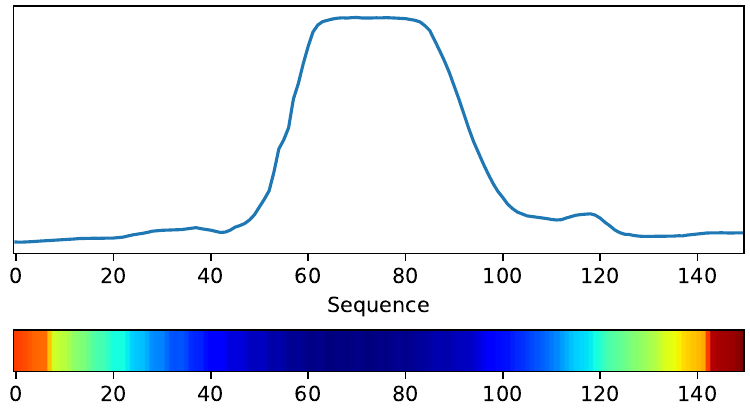}}%
		\label{fig:dev8dR1}%
	}\qquad
 	\subfigure[Input and T0 CAM output for class 1]{%
 		{\includegraphics[width=0.30\textwidth, trim= 1 1 1 1,clip, keepaspectratio]{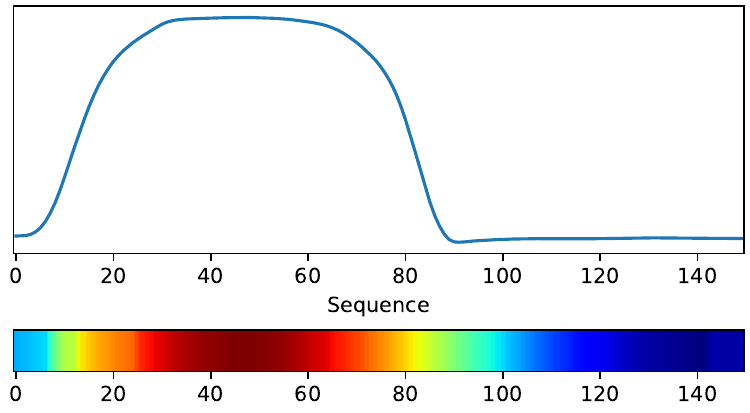}}%
 		\label{fig:dev9dR2}%
 	}
 	\\
	\subfigure[T1 CAM output for class 0]{%
		{\includegraphics[width=.30\linewidth, trim= 1 1 1 158,clip, keepaspectratio]{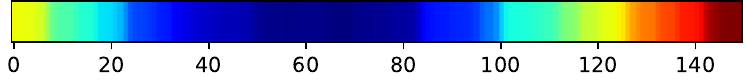}}%
		\label{fig:dev8dR2}%
	}\qquad
 	\subfigure[T1 CAM output for class 1]{%
 		{\includegraphics[width=0.30\textwidth, trim= 1 1 1 158,clip, keepaspectratio]{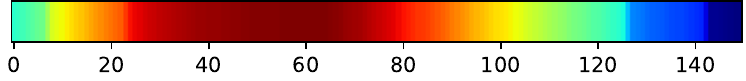}}%
 		\label{fig:dev9dR1}%
 	}
 	\\
	\subfigure[T$\protect_{\beta}$ CAM output for class 0]{%
		{\includegraphics[width=.30\linewidth, trim= 1 1 1 158,clip, keepaspectratio]{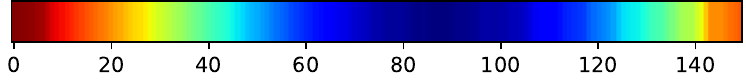}}%
		\label{fig:dev8dR3}%
	}\qquad
 	\subfigure[T$\protect_{\beta}$ CAM output for class 1]{%
 		{\includegraphics[width=0.30\textwidth, trim= 1 1 1 158,clip, keepaspectratio]{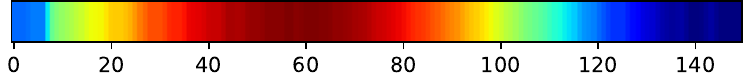}}%
 		\label{fig:gpo_cam}%
 	}
	\caption{GunPointOldVersusYoung dataset}
	\label{fig:gpo}
\end{figure*}

\begin{figure*}[t]
	\centering
	
	\subfigure[Input and T0 CAM output for class 0]{%
		{\includegraphics[width=.30\linewidth, trim= 1 1 1 1,clip, keepaspectratio]{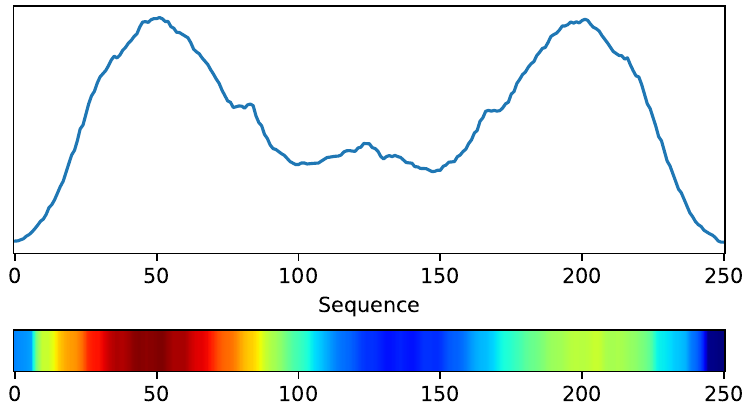}}%
		\label{fig:CAM_db_ArrowHead_cls_0_config_T0}%
	}\qquad
 	\subfigure[Input and T0 CAM output for class 1]{%
 		{\includegraphics[width=0.30\textwidth, trim= 1 1 1 1,clip, keepaspectratio]{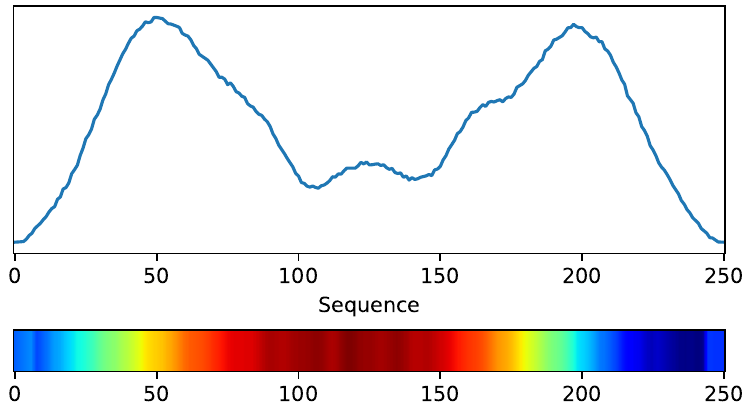}}%
 		\label{fig:CAM_db_ArrowHead_cls_1_config_T0}%
 	}\qquad
 	\subfigure[Input and T0 CAM output for class 2]{%
 		{\includegraphics[width=0.30\textwidth, trim= 1 1 1 1,clip, keepaspectratio]{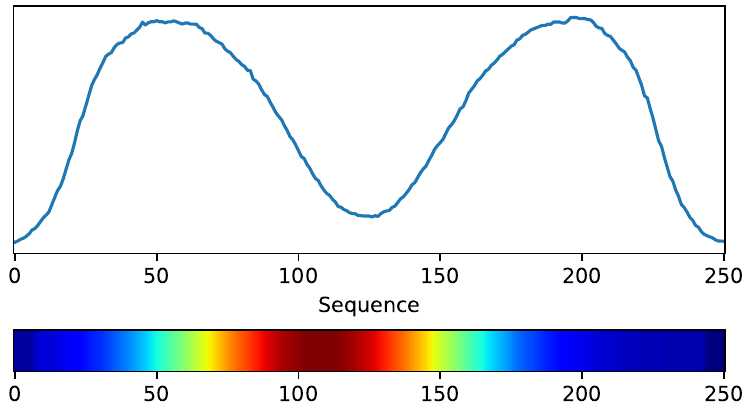}}%
 		\label{fig:CAM_db_ArrowHead_cls_2_config_T0}%
 	}
 	\\
	\subfigure[T1 CAM output for class 0]{%
		{\includegraphics[width=.30\linewidth, trim= 1 1 1 158,clip, keepaspectratio]{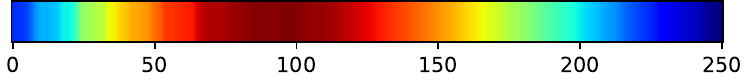}}%
		\label{fig:CAM_db_ArrowHead_cls_0_config_T1}%
	}\qquad
 	\subfigure[T1 CAM output for class 1]{%
 		{\includegraphics[width=0.30\textwidth, trim= 1 1 1 158,clip, keepaspectratio]{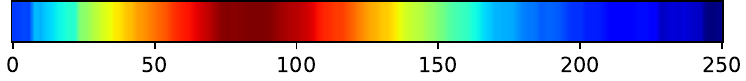}}%
 		\label{fig:CAM_db_ArrowHead_cls_1_config_T1}%
 	}\qquad
 	\subfigure[T1 CAM output for class 2]{%
 		{\includegraphics[width=0.30\textwidth, trim= 1 1 1 158,clip, keepaspectratio]{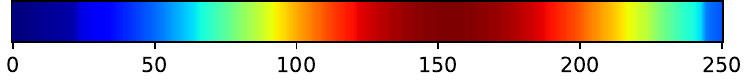}}%
 		\label{fig:CAM_db_ArrowHead_cls_2_config_T1}%
 	}
 	\\
	\subfigure[T$\protect_{\beta}$ CAM output for class 0]{%
		{\includegraphics[width=.30\linewidth, trim= 1 1 1 158,clip, keepaspectratio]{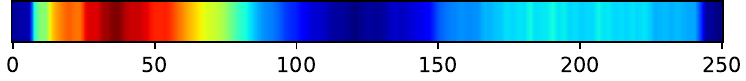}}%
		\label{fig:CAM_db_ArrowHead_cls_0_config_T2}%
	}\qquad
 	\subfigure[T$\protect_{\beta}$ CAM output for class 1]{%
 		{\includegraphics[width=0.30\textwidth, trim= 1 1 1 158,clip, keepaspectratio]{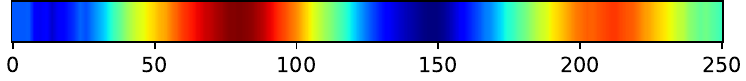}}%
 		\label{fig:CAM_db_ArrowHead_cls_1_config_T2}%
 	}\qquad
 	\subfigure[T$\protect_{\beta}$ CAM output for class 2]{%
 		{\includegraphics[width=0.30\textwidth, trim= 1 1 1 158,clip, keepaspectratio]{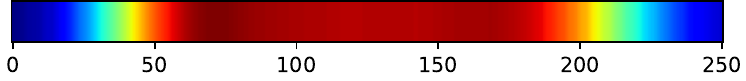}}%
 		\label{fig:CAM_db_ArrowHead_cls_2_config_T2}%
 	}
	\caption{ArrowHead dataset}
	\label{fig:arrow_head}
\end{figure*}

In this section, we examine the effects of altering the kernel and channel sizes. In order to accomplish this, the ATCN model $T_{\beta}$ is defined as having a channel size that is twice as large as T1 per layer, however, the kernel size is half the size. Due to this configuration, $T_{\beta}$ has greater model complexity than T1, both in terms of FLOPS and model size, while still achieving the same accuracy. Table \ref{tab:t_beta} summarizes the $T_{\beta}$ network configuration, FLOPs, and the number of parameters. It can be seen that $T_{\beta}$ has the same $\boldsymbol{D}$, and $\boldsymbol{R}$ as T1, but $\boldsymbol{C^{\beta}_{out}}=2\times\boldsymbol{C^{T1}_{out}}$ and the kernel size per block is half. Despite a greater model complexity than T1, both FLOPs and model size, the results depicted in Fig. \ref{fig:t1_tbeta} show that overall T1 and $T_{\beta}$ achieve very similar accuracy. As an aid to understanding this behavior, we depict the inputs and Class Activation Mapping (CAM) \cite{CAM} of two benchmarks in Fig. \ref{fig:gpo} and Fig. \ref{fig:arrow_head}. When the model correctly classified the input signal, CAMs are calculated by multiplying the input of global average pooling by the weight matrices of the correct class index. 

\input{Tables/t_beta}

The activation heatmaps for class 0, shown in Fig.~\ref{fig:gpo}, have the lowest value around the input signal magnitude. Consequently, for class 0, the probability of output approaches zero. However, we can observe the activation heatmaps for class 1 have the highest value around the signal magnitude, which leads to the output probability being one. For multiclass classification problems depicted in Fig.~\ref{fig:arrow_head}, we can see models activate based on the perceived nuances of signal shapes. For instance, the T0 model, Fig.~\ref{fig:CAM_db_ArrowHead_cls_0_config_T0}, classifies the input as class 0 based on the form observed in sequence $\sim$10 to $\sim$80, as class 1 based on unique transition observed in the middle of the sequence, and as class 2 based on the curve recognized in $\sim$70 to $\sim$150. In respect to $T_{\beta}$,  this model shows a coarse-grained transition, note sequence $\sim$20 to $\sim$60 in Fig.~\ref{fig:gpo_cam} and $\sim$150 to $\sim$240 in Fig.~\ref{fig:CAM_db_ArrowHead_cls_0_config_T2}. This indicates that $T_{\beta}$ has a lower receptive field compared to T1. As a result of the higher receptive field of T1, the model is able to predict the classes more precisely, although it has less model complexity in both forms of FLOPS and the number of parameters.

%% file: Tables/runTime.tex
\begin{table}[htbp]
  \centering
  \caption{Resource utilization and inference time of Cortex-M7}
 \begin{tabular}{lcccc}
    \toprule
    \multicolumn{1}{c}{\multirow{2}[4]{*}{\textbf{Description}}} & \multicolumn{4}{c}{\textbf{Models}} \\
\cmidrule{2-5}          & \textbf{T0} & \textbf{T1} & \textbf{MiniRock} & \textbf{InceptionTime} \\
    \midrule
    \textbf{M7 Ram utilization (\%)} & 48.89 & 56.39 & -     & - \\
    \textbf{M7 Flash utilization (\%)} & 14.13 & 15.89 &  -     & - \\
    \textbf{M7 inference time (ms)} & 165   & 210   &  -     & - \\
    \bottomrule
    \end{tabular}%
  \label{tab:run_time}%
\end{table}%

%% file: Tables/pareto.tex
\begin{table}[htbp]
  \centering
  \caption{The performance of three T0, T1 and MiniRocket in contribution of 144 Pareto solutions.}
    \begin{tabular}{ccccccc}
    \toprule
    \multicolumn{1}{c}{\multirow{3}[6]{*}{\textbf{Description}}} & \multicolumn{6}{c}{\textbf{Models}} \\
\cmidrule{2-7}    \multicolumn{1}{c}{} & \multicolumn{2}{c}{\textbf{T0}} & \multicolumn{2}{c}{\textbf{T1}} & \multicolumn{2}{c}{\textbf{MiniRocket}} \\
\cmidrule{2-7}    \multicolumn{1}{c}{} & \textbf{Total} & \textbf{Unique} & \textbf{Total} & \textbf{Unique} & \textbf{Total} & \textbf{Unique} \\
    \midrule
    \textbf{Contribution} & \textbf{66}    & \textbf{15}    & 39    & 0     & 39    & 2 \\
    \bottomrule
    \end{tabular}
  \label{tab:pareto}%
\end{table}%

%% file: Tables/t_beta.tex
\begin{table}[htbp]
  \centering
  \caption{Model configuration and accuracy performance of $T_{\beta}$}
  \begin{adjustbox}{width=1.0\linewidth,center}
    \begin{tabular}{lcccccc}
    \toprule    
    \multicolumn{1}{c}{\multirow{2}[4]{*}{\textbf{Model}}} & \multicolumn{6}{c}{\textbf{Parameters}} \\
    \cmidrule{2-7}          & $\boldsymbol{C_{out}}$ & $\boldsymbol{D}$ & \textbf{$\boldsymbol{K}$} & $\boldsymbol{R}$ & \textbf{FLOPs} & \textbf{Params\#}\\ 
    \midrule
    $T_{\beta}$    & $[64, 32, 32, 16, 16, 32, 32, 64]$ & $[1, 2, 2, 4, 4, 6, 6, 8]$ & $[32, 16, 16, 8, 8, 4, 4, 2]$ & $[\frac{1}{2},  1,   1,  1,  1,  1,  1, 1]$ & 7,303,136 & 86,240  
    \\
    \bottomrule
    \end{tabular}%
    \end{adjustbox}
  \label{tab:t_beta}%
\end{table}%

%% file: Tex/Conclusion.tex
\section{Conclusion}\label{sec:conl}
This article proposed ATCN which is a novel family of networks for real-time processing of time-series on embedded and edge devices. In order to reduce the number of MAC operations and model size, we introduced STCB as a main computational block. STCB blocks are able to be sequenced in a variety of configurations to build scalable ATCNs. We also presented a framework, called \textit{ATCN Model Synthesizer}, to build different ATCN models. The result of \textit{ATCN Model Synthesizer} is a family of compact networks with formalized hyper-parameters that allow the model architecture to be configurable and adjusted based on the application requirements. Through the use of model synthesizer and ATCN reconfigurability, we have developed two fast while accurate models, T0, T1 which can be executed on ARM Cortex-M7 microcontrollers. The experimental results over 70 benchmarks of 2018 UCR time classification dataset indicate that the T0 configuration can reduce the MACs and model size by 102.38$\times$ and 16.84$\times$ over InceptionTime, respectively. T1 also has a 73.59$\times$ reduction in MACs, and a 14.23$\times$ reduction in model size over InceptionTime. In addition, we compared the performance of T0 and T1 against MiniRocket as a benchmark for a fast and accurate non-deep learning time series classifier. According to our results, the T0 model outperforms MiniRocket across 15 benchmarks and provides 66 optimal solutions out of 144 Pareto points, while MiniRocket only contributes 39 points.

%% file: Tex/Appendix.tex
\appendix
\section{Bar graph Comparison of Models}\label{sec:app}
\begin{figure}[h]
    \centering
    \includegraphics[width=.86\textwidth, trim= 1 1 1 1,clip, keepaspectratio]{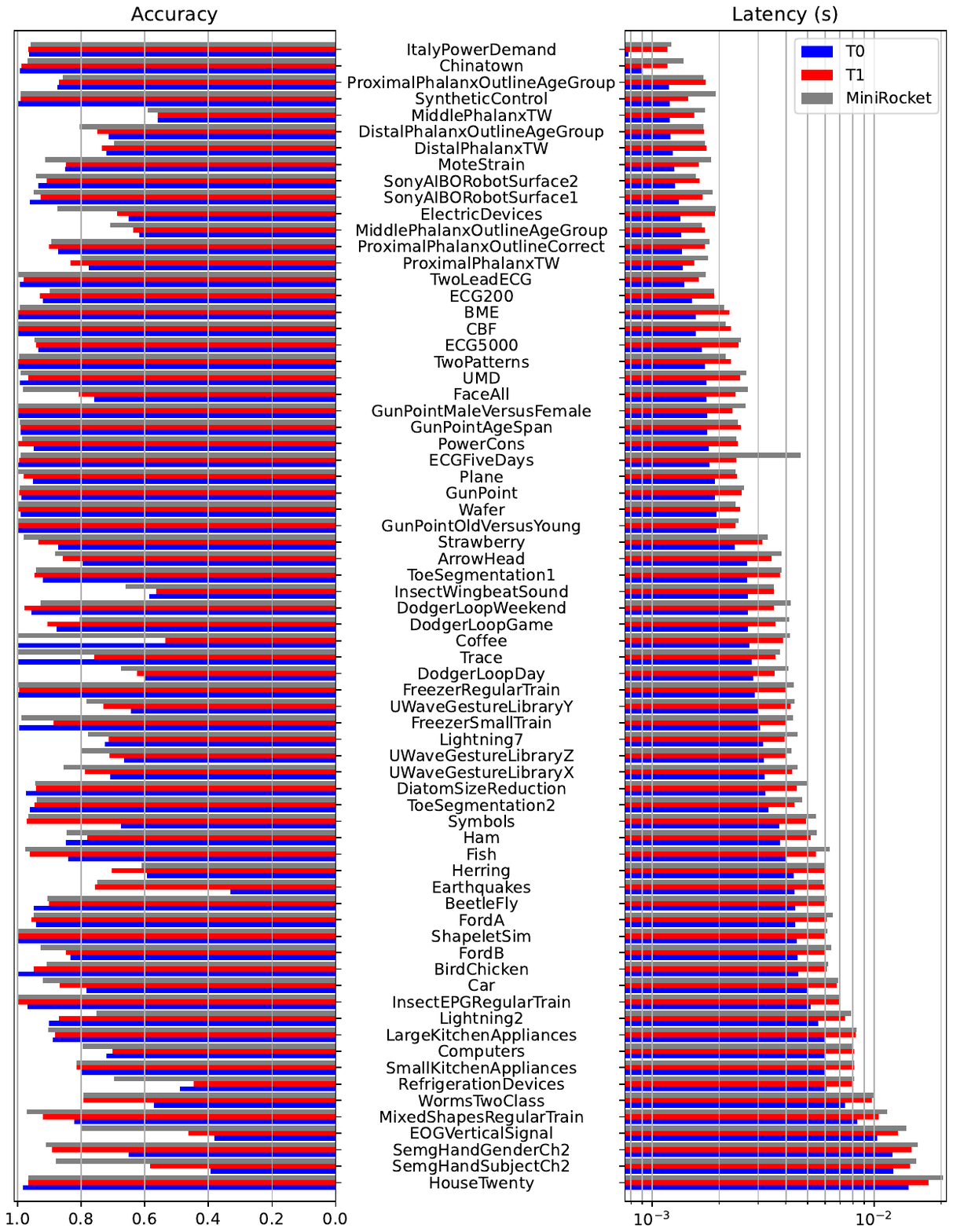}    
    \vspace{-15pt}
    \caption{Comparison of accuracy and latency of three models on a Cortex-A57 processor. Results are sorted by T0 latency.}
    \label{fig:dt_results}
\end{figure}

Fig. \ref{fig:dt_results} compares the accuracy and latency of the T0, T1, and MiniRocket models across all seventy benchmarks. As explained in Section 5, latency values are obtained by running benchmarks on Cortex-A57. In Fig. \ref{fig:dt_results}, we sorted the results based on T0 latency.

